**Large Language Model Displays Emergent Ability to Interpret Novel Literary Metaphors**

Nicholas Ichien[a,1]     Dušan Stamenković[b.]     Keith J. Holyoak[c]

[a]Department of Psychology, University of Pennsylvania, PA 19104
[b]School of Culture and Education, Södertörn University, Stockholm, Sweden
[c]Department of Psychology, University of California, Los Angeles, CA 90095-1563

[1]Corresponding author

Nicholas Ichien
Department of Psychology
University of Pennsylvania
3720 Walnut Street
Phildelphia, PA, 19104
nichien@sas.upenn.edu

Dušan Stamenković
School of Culture and Education
Södertörn University
Stockholm, Sweden
dusan.stamenkovic@sh.se

Keith J. Holyoak
Department of Psychology
University of California, Los Angeles
405 Hilgard Ave.
Los Angeles, CA 90095-1563
kholyoak@g.ucla.edu

**Abstract**: Despite the exceptional performance of large language models (LLMs) on a wide range of tasks involving natural language processing and reasoning, there has been sharp disagreement as to whether their abilities extend to more creative human abilities. A core example is the interpretation of novel metaphors. Given the enormous and non-curated text corpora used to train LLMs, a serious obstacle to designing tests is the need to obtain novel yet high-quality metaphors that are unlikely to have been included in the training data. Here we assessed the ability of GPT-4, a state-of-the-art large language model, to provide natural-language interpretations of a recent AI benchmark (Fig-QA dataset), novel literary metaphors drawn from Serbian poetry and translated into English, and entire novel English poems. GPT-4 outperformed previous AI models on the Fig-QA dataset. For metaphors drawn from Serbian poetry, human judges—blind to the fact that an AI model was involved—rated metaphor interpretations generated by GPT-4 as superior to those provided by a group of college students. In interpreting reversed metaphors, GPT-4, as well as humans, exhibited signs of sensitivity to the Gricean cooperative principle. In addition, for several novel English poems GPT-4 produced interpretations that were rated as excellent or good by a human literary critic. These results indicate that LLMs such as GPT-4 have acquired an emergent ability to interpret literary metaphors, including those embedded in novel poems.

The poet Robert Frost asserted, "The richest accumulation of the ages is the noble metaphors we have rolled up" (Frost, 1931, p. 108). The world's literature (Holyoak, 2019; Lakoff & Turner, 1989; Steen, 1994), as well as everyday speech (Lakoff & Johnson, 1980), is replete with non-literal comparisons of things that are on the face of it unlike each other, e.g., "'Hope' is the thing with feathers —That perches in the soul" (Emily Dickinson). The ability to create and interpret novel metaphors is considered one of the pinnacles of human cognitive abilities, extending literal language and perhaps involving sophisticated analogical reasoning (Bowdle & Gentner, 2005; Holyoak & Stamenković, 2018). If artificial intelligence (AI) aims to ultimately reach or exceed human cognitive abilities, then models of natural language processing and general intelligence will need to acquire the ability to interpret (and perhaps create) novel metaphors.

**1. Metaphor interpretation by large language models**

The advent of large language models (LLMs) has triggered intense interest in whether these new AI models are in fact approaching human-level abilities in language understanding (DiStefano et al., 2023; Köbis & Mossink, 2021; Mahowald et al., 2023; McClelland et al., 2020) and various forms of reasoning (Binz & Schulz, 2023; Chan et al., 2022; Dasgupta et al., 2022; Srivastava et al., 2022; Wei et al., 2022), including analogy (Webb et al., 2023), with vigorous debate between critics (Chomsky et al., 2023) and advocates (Piantadosi, 2023). Given the enormous and non-curated text corpora on which LLMs have been trained, these models have certainly had ample opportunity to mine the metaphors that humans have already formed and planted in texts. More generally, recent reports suggest that LLMs have been exposed to some or all standard AI benchmarks for cognitive tasks, rendering these benchmarks suspect for distinguishing deeper cognitive processing from memorization (Bubeck et al., 2023). It would not be surprising to find that LLMs succeed on tasks involving linguistic expressions contained in their training data, including metaphors. A serious test of a model's ability to deal with novel metaphors requires challenging it with metaphors that are both novel and *apt* (i.e., metaphors in which the source is perceived as providing a unique and accurate description of the target). Unfortunately, detailed information about the training data for LLMs is at present unavailable, making it difficult to determine definitively whether any given metaphor is truly novel for the system. All conventional metaphors, e.g., "Life is a journey," are certain to have been included in the text corpora used as the training set. Creating novel metaphors that have never been uploaded to the internet is extremely challenging. Moreover, there is evidence that new metaphors generated by psychologists for experimental purposes are typically weak in aptness (Jones & Estes, 2006). In general, those metaphors that become conventional are those people have considered to be particularly apt.

## 2. Overview of the paper

Here we test GPT-4, a state-of-the-art LLM, on its ability to generate natural-language interpretations of literary metaphors that passed tests assessing their novelty to the model. Computational analyses have shown that literary metaphors are distinguished by the qualities of high surprisal (a statistical measure of the unexpectedness of words), relative dissimilarity of source and target concepts, the combination of concrete words with relatively complex grammar and high lexical diversity, and extra difficulty (for people) in comprehending the metaphorical meaning (Baggio, 2018; Jacobs & Kinder, 2017, 2018). Studies of individual differences in cognitive abilities have shown that crystalized intelligence (closely linked to verbal ability) impacts comprehension of both conventional and literary metaphors; fluid intelligence (on which analogical reasoning depends heavily) plays a greater role for more complex literary metaphors (especially when presented in isolation without a supportive verbal context) (Stamenković et al., 2020, 2023; Stamenković, Ichien, et al., 2019). Novel literary metaphors thus pose the most challenging test of metaphor comprehension in humans and perhaps AI models.

We report three studies that examine the ability of GPT-4 to identify or generate interpretations of metaphors. Study 1 provides an extensive comparison between GPT-4 and previous state-of-the-art language models on a two-alternative force-choice metaphor comprehension task based on the Fig-QA dataset developed by Liu et al. (2022)). Study 2 goes beyond previous work evaluating large language models on their metaphor comprehension ability (Liu et al., 2022; Prystawski et al., 2023) by testing stimuli drawn from poetry, thus consisting of much more naturalistic literary language. Finally, Study 3 assesses the quality of GPT-4's interpretations of poems written by published poets (but not available for GPT-4's training), as evaluated by a professional literary critic. The metaphors examined in Studies 2-3 contrast with the highly simplistic and formally stereotyped materials designed for use in previous psychology experiments and tests of AI models.

## 3. Interpreting metaphors in Fig-QA dataset

*3.1 Background*

Recent work has evaluated a number of LLMs (GPT-2, GPT-neo, GPT-3, BERT, and RoBERTa) on their ability to generate interpretations of metaphors (Liu et al., 2022). In a forced-choice paradigm, models and humans were given a metaphor from the Fig-QA dataset (e.g., "The clothing has the smell of a crowded gym") and selected one of two literal rephrases of that metaphor (e.g., "The clothing smells pleasant" vs. "The clothing smells awful"). Humans made their selection directly, whereas models did so implicitly by determining which rephrase yielded a higher value on some metric of sentence plausibility (e.g., log probability for GPT models) when plugged into the following form: "<METAPHOR>, that is to say, <REPHRASE>" (e.g., "The clothing has the smell of a crowded gym, that is to say, the clothing smells awful").

Across the 1,146 metaphors in the test set of Liu et al.'s Fig-QA dataset (https://github.com/nightingal3/Fig-QA; leaderboard can be found at: https://explainaboard.inspiredco.ai/leaderboards?dataset=fig_qa), human participants did very well, achieving 94.42% accuracy (95.39% after excluding any problems for which participants reported lack of confidence in their response). Of the models that Liu et al. evaluated, GPT-3 Davinci (Brown et al., 2020; Radford et al., 2019), the largest model (175B parameters, trained on 45TB of text) achieved the best zero-shot performance (68.41%) but still fell well-short of human-level responding. RoBERTa came close to human-level responding (90.32%), but only after it was fine-tuned on 8,016 similar metaphors.

3.2 Method and results

We tested GPT-4 zero-shot by directly prompting the model with natural language text using the following prompt form: "Which of the following is a better interpretation of the following expression: <METAPHOR>. A: <REPHRASE 1> Or B: <REPHRASE 2>?". The model provided an unambiguous response (e.g., "B. The clothing smells awful") most of the time, but on 3.35% of questions, it responded with some variant of "Neither A nor B is a direct interpretation of <METAPHOR>." Counting the latter responses as incorrect, along with those for which the model responded with the wrong interpretation, GPT-4 outperformed both GPT-3 Davinci zero-shot and the fine-tuned RoBERTa, achieving close to human-level performance (91.27%). Thus GPT-4 far outperformed other LLMs in zero-shot performance, and even bested a fine-tuned variant of RoBERTa. GPT-4 is considerably larger (1.8 trillion parameters, trained on 1 PB of text and images) than GPT-3 Davinci, the largest model Liu et al. tested (OpenAI, 2023). GPT-4 was trained not only on text, but also computer code and images. It also incorporated reinforcement learning with human feedback (RLHF), which served to further align its text output with that generated by humans and rated by them as highly appropriate.

The metaphors used in the Fig-QA dataset were all crowdsourced from MTurk workers and the dataset was not published until June 23, 2022—later than September 2021, the latest date that the data used to train GPT-4 were collected (OpenAI, 2023). It is therefore unlikely that model performance solely reflects GPT-4's ability to reproduce its memorized training data. However, metaphors generated by MTurk workers are unlikely to be entirely original. Those included in the Fig-QA dataset were clearly easy for humans to reliably interpret, as they achieved nearly ceiling performance. Accordingly, in Studies 2-3 we examined literary metaphors that were unlikely to have been included in datasets used to train GPT-4.

**4. Interpreting metaphors drawn from Serbian poetry**

*4.1. Method*

Study 2 was performed using materials designed to minimize the possibility that GPT-4 had been trained on our test metaphors and could produce interpretations based on memory for these specific examples. Specifically, we selected metaphors that originated in Serbian poems and had been translated into English. The original metaphors were rated as highly apt by native Serbian speakers, but were not widely known to them; we assessed and then controlled for the familiarity of the English translations to English-speaking participants and to GPT-4.

*4.1.1 Selection of metaphors from Serbian poetry*

The test set included 55 literary metaphors drawn from Serbian poetry and normed on several properties (Milenković et al., In review; Stamenković, Milenković, et al., 2019). The norming studies (primarily following methods used previously (Katz et al., 1988)) included metaphors chosen by a literary expert from over 65 nineteenth and twentieth-century poems written by various Serbian poets, including Branko Radičević, Laza Kostić, Vojislav Ilić, Đura Jakšić, Desanka Maksimović, Vladislav Petković Dis, and Branko Miljković. The poems selected for the norming study aimed to represent a wide range of poetic movements and styles across these two centuries. The expert had the task of extracting all metaphorical expressions from these poems. These were then grouped, with all similar/duplicate metaphors counted as one. Subsequently, all metaphors were transformed into *<nominal> is <nominal>* format, resulting in the finalized list of 55 items.

In the first norming study, 235 Serbian-speaking participants rated these 55 metaphors for quality, metaphoricity, aptness, familiarity, comprehensibility, and source-target similarity using a 7-point Likert scale (min = 1, max = 7). The inter-scale correlations for Serbian poetic metaphors were reliable for most dimensions (Stamenković, Milenković, et al., 2019) (ranging from .77 between aptness and quality, and between quality and source-target similarity, to .27 between familiarity and metaphoricity). The only nonsignificant correlation was that between source-target similarity and metaphoricity (.17). These 55 literary metaphors were then translated into English by two translators and the translations were verified by a third. This new list was subjected to two further norming studies (Milenković et al., In review) in which a combined total of 252 (186 for the full set and 66 for a shorter set) English-speaking Serbian participants rated the set for quality, metaphoricity, aptness, and familiarity (see Table S1 in Section 1 of Supplemental Online Information). These Serbian translations were selected to minimize the likelihood that their English translations appeared in GPT-4's training data (see below, "Assessment of Novelty of Serbian Metaphors to GPT-4").

*4.1.2 Assessment of novelty of Serbian metaphors to GPT-4*

We performed several tests to evaluate whether GPT-4 was familiar with the metaphors used in our experiment. First, we probed it with questions to assess its knowledge of one paper (Stamenković, Milenković, et al., 2019) and an online supplement to another paper (Stamenković et al., 2023), in which some of these metaphors were discussed. Neither of these publications included the metaphor interpretations that GPT-4 generated in our tests. GPT-4 recognized one paper (Stamenković, Milenković, et al., 2019) and summarized its main points, but was unable to report any of the metaphors in it. When asked for examples, it provided several unrelated metaphorical expressions that did not appear in the study ("hallucinations"). The later paper (Stamenković et al., 2023) was published in March 2023, making it too recent to have been included in GPT-4's training data; the model did not recognize its online supplement at all.

Second, we considered the possibility that the Serbian-language version of GPT-4 might be familiar with the original Serbian poems from which our materials were derived. We therefore performed an additional test conducted entirely in Serbian using the Serbian-language version of GPT-4. We first provided the model with a list of the 65 poems from which our metaphor stimuli were drawn and adapted, with the query (translated from Serbian): "This is a list of poems and their respective authors. Tell me if you recognize any of them." GPT-4's responses were uniformly vague, so we added a follow-up prompt: "Look at the list of poems I have provided and tell me which of them you could analyze if I asked you to do so." Out of the 65 poems, GPT-4 indicated it could analyze 14. We then assessed the system's knowledge of these 14 poems using a "fill-in-the-gap" exercise, using the query: "Could I provide poems with blanks, where your task would be to fill in these blanks with the missing parts?" In this test, we provided the system with each poem's title and author, along with the poem itself, omitting the lines that contained our selected metaphors (in their original form), and we prompted GPT-4 to fill in the missing lines. Despite being given the above details about each poem, the system was unable to generate the omitted lines for any of the poems. Instead, it generated its own replacements for the missing lines, trying to imitate the style of the poet and the poem, and sometimes explicitly stating that the poem is not available in its current database. We thus found no evidence that GPT-4 had been exposed to the Serbian versions of the metaphors used in our test.

Third, we considered the possibility that at least some of the translated Serbian metaphors had been generated previously in English and uploaded to the internet. In the absence of direct access to GPT-4's training corpus, we quote-searched each metaphor using Google's search engine. Of the 55 metaphors drawn from Serbian poetry, we found a match for 19 of their English translations published online before 2022 (34.5%). For example, we found the expression "Love is radiance" in the poem "Love" by Miklos Zoltai, published online in March 2016: "Love is radiance, lover is perfumed of your self-knowledge, love is the essence of your existence." We coded each translated metaphor as having some online match or not. This variable was included as a predictor in our analyses to assess whether the previous online publication (and hence potential inclusion in GPT-4's training corpus) influenced ratings of GPT-4's metaphor interpretations by human judges. (See interaction term *interpretation source x online match (yes vs. no)* in analyses of data presented in Figure 1 below.)

*4.1.3 Human experiment and comparison with GPT-4*

In order to evaluate GPT-4's ability to generate high-quality interpretations of metaphors, we compared its interpretations with that of 39 undergraduate psychology students at the University of California, Los Angeles (UCLA), who completed our task for course credit (approved, including informed consent procedures, by the UCLA Office of the Human Research

Protection Program). In order to elicit responses from human participants and GPT-4, we plugged each metaphor into the following prompt: "Please provide an interpretation for the following expression: <SENTENCE>", where the sentence stated the metaphor. Because GPT-4 tended to provide longer interpretations, we prompted it to provide a "short sentence-long interpretation" to elicit responses that were relatively succinct and of similar length to those generated by human participants. Importantly, the experimental task for both human participants and for GPT-4 omitted any mention of the term 'metaphor', opting instead for the more neutral term 'expression' as used above. Thus, no overt cue indicated that the task had anything to do with metaphor.

In order to provide a qualitative assessment of GPT-4's text interpretations, we also assessed the extent that both human- and model-generated text interpretations followed the Gricean cooperative principle, which implies that inapt or poor-quality metaphors may be reinterpreted as expressing a more apt and informative comparison than what they literally say (Chiappe et al., 2003; Grice, 1975). To do so, we elicited interpretations of literary metaphors both in their canonical (apt) form (e.g., "Love is radiance") as well as in their reversed (inapt) form (e.g., "Radiance is love"; see Table 1 for additional examples).

Human participants provided interpretations for 55 metaphor items, each in either its canonical form (e.g., "Love is radiance") or reversed form with source and target switched (e.g., "Radiance is love"). Which form a metaphor took was randomized across items for each participant. Across all unique items (110 total with 55 canonical form and 55 reversed), we collected an average of 19.54 human responses for each item (range = [12, 27]). GPT-4 provided an interpretation of all 110 metaphor items. We used GPT-4's chat interface (in late March 2023) to collect each response after starting a new chat window. This procedure prevented the system from conditioning its response on previously-seen metaphor items, so that its performance was zero-shot.

Table 1. Example of GPT-4's interpretations of metaphors presented in their canonical direction (left columns) and in their reversed direction (right columns).

| Canonical | | Reversed | |
| --- | --- | --- | --- |
| Metaphor | GPT-4 interpretation | Metaphor | GPT-4 interpretation |
| A waterfall is a wild, unbridled horse. | A waterfall symbolizes the untamed, powerful energy of nature, much like the spirit of a wild, unbridled horse. | A wild, unbridled horse is a waterfall. | A wild, unbridled horse is a waterfall suggests that unrestrained energy and freedom can possess a raw, captivating beauty. |
| A cemetery is a sad theatre. | A cemetery is a melancholic stage where the final acts of life are commemorated, and loved ones gather to mourn and remember the departed. | A sad theatre is a cemetery. | A sad theatre is a cemetery conveys that a place once filled with emotion and expression can become desolate and lifeless when devoid of its original purpose. |
| Love is radiance. | Love is the glowing warmth that illuminates and uplifts our lives. | Radiance is love. | Radiance is love suggests that the emotion of love has the power to illuminate and transform our lives, shining with warmth and positivity. |

After eliciting interpretations from both humans and GPT-4, three undergraduate raters who were naïve to presence of nonhuman, model-generated text scored each interpretation on a 0-2-point scale. Directions for their scoring are provided in Section 2 of Supplemental Online Information, and rater-generated examples are available on this paper's OSF page: https://osf.io/jcg3f/?view_only=eed36a8a1b9e410aaef1926d020300f9. Briefly, raters were instructed to assign interpretations a score of 2 if they attributed properties of the source to the target that described the target aptly. Interpretations were to receive a 1 if they attributed certain properties to the target that seemed less appropriate, especially given the source. Finally, interpretations received a 0 if they were blank or failed to express any sort of comparison. For each metaphor, raters were given a set of interpretations that had been shuffled so as to mask any common author of interpretations across items.

*4.2 Results*

To analyze our data, we fit a cumulative link mixed model to trial-level interpretation scores (0, 1, 2 treated as an ordinal variable), using the *clmm* function from version 2022.11.16 of the ordinal R package (Christiansen, 2022) in R version 4.3.1 (R Core Team, 2021). We defined a full model including *participant*, *judge*, and *metaphor* as random intercept effects, and including two two-way interaction terms as fixed effects: *interpretation source* (*GPT-4* vs. *human*) x *metaphor form* (*canonical* vs. *reversed*) and *interpretation source x online match (yes vs. no)*. Figure 1 shows metaphor score data, broken down according to metaphor form, with GPT-4 coded as one participant.

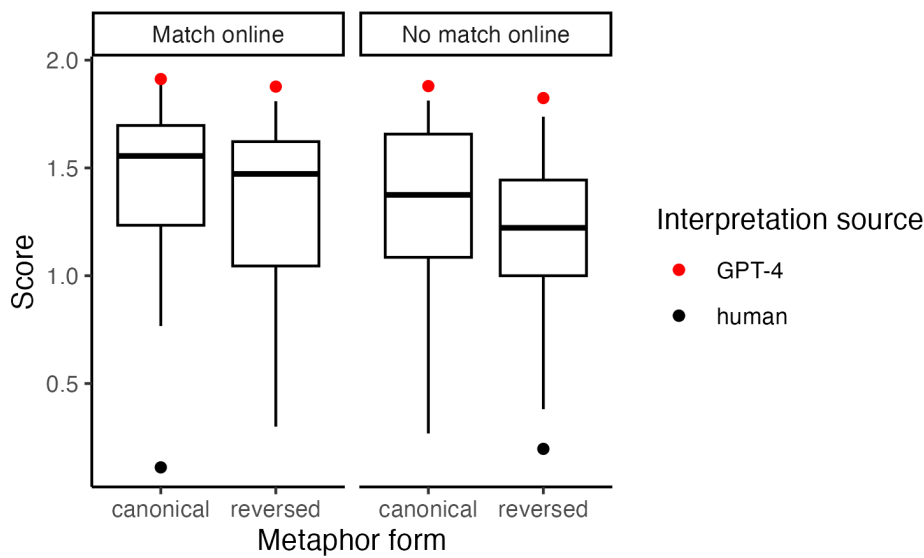

*Figure 1*. Metaphor interpretation scores (averaged across judges) broken down by metaphor form ("canonical" vs. "reversed") and whether a metaphor's English translation was found online published before 2022 ("Match online" vs. "No match online"). Human data are represented as a boxplot and GPT-4 performance is represented by red points.

All statistical tests involve comparing the full model described above with an ablated model that lacks a particular predictor of interest (e.g., coding source) but that is otherwise equivalent to the full model. We then use likelihood ratio tests to evaluate whether or not the ablated model yields increased prediction error compared to the full model: To the extent that it does, the omitted predictor was crucial to model performance and was thus an important predictor of the modeled data.

Removing the *interpretation source x metaphor form* interaction term did not increase model prediction error, $\Delta$AIC = 2, $\chi^2(1) = .01$, $p = .92$, and neither did removing the *interpretation source x online match* interaction term, $\Delta$AIC = 2, $\chi^2(1) = .05$, $p = .81$, nor did jointly removing both interaction terms, $\Delta$AIC = 4, $\chi^2(1) = .07$, $p = .97$. We thus did not detect any performance difference between GPT-4 and human participants that varied as a function of metaphor form or as a function of whether or not a given metaphor was found online pre-2022 (and thus possibly in GPT-4's training corpus). Inspecting the fit model that omitted these interaction terms, we found main effects for both interpretation source ($\beta = .194$, $z = 2.01$, $p = .045$), metaphor form ($\beta = .383$, $z = 7.50$, $p < .001$), and online match ($\beta = .379$, $z = 2.29$, $p = .022$). As is evident from Figure 1, GPT-4 outscored all human participants. The AI system and humans were both affected by the metaphor form, such that interpretations of metaphors in the reversed form received lower scores than those in the canonical form. Both the model and humans were also consistently affected by a metaphor's presence online. A metaphor being published online increases the probability that both GPT-4 and human reasoners may have encountered the expression prior to our experiment, providing a rough indication of its familiarity. Previous work has shown that more familiar metaphors tend to be comprehended more easily than those that are less familiar (Blasko & Connine, 1993; Stamenković, Ichien, et al., 2019).

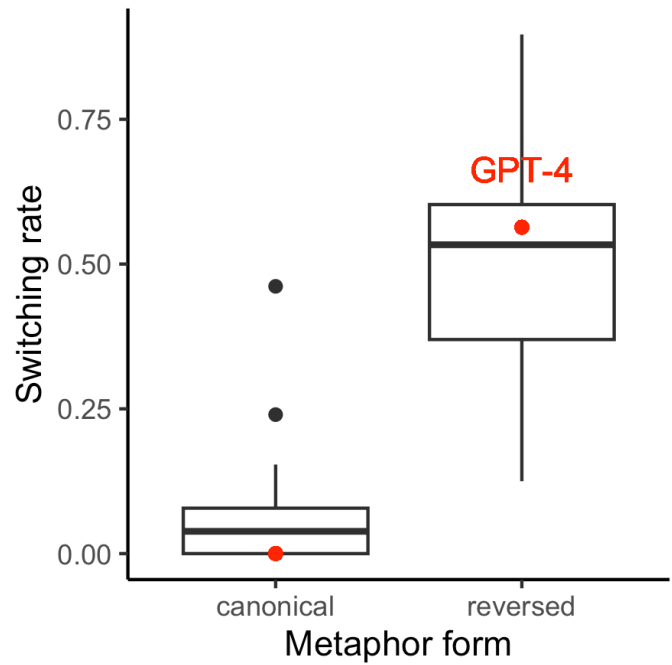

*Figure 2*. Metaphor switching rates (proportion of times an interpretation described the source of a metaphor, rather than its target) broken down by metaphor form (canonical vs. reversed). Human data are represented as a boxplot and GPT-4 performance is represented by red points.

When the metaphors were presented in the reversed (non-canonical) order of source and target (e.g., canonical: "A man is a butterfly" vs. reversed: "A butterfly is a man"), human participants often gave interpretations that restored the canonical order (50% of responses; "A man goes through many phases of growth in order to reach his full potential."). More generally, we coded the rate at which roles were switched for both canonical and reversed metaphors. As shown in Figure 2, switching was very rare (as would be expected) when the metaphor was presented in canonical form but was very common when the metaphor was reversed (in which case switching restored the canonical interpretation). Restoring the canonical meaning of a reversed metaphor is consistent with previous findings concerning how people interpret reversed metaphor (Chiappe et al., 2003). This propensity likely reflects the Gricean cooperative principle, according to which people seek effective communication (Grice, 1975). Remarkably, GPT-4 gave interpretations of reversed metaphors that restored their canonical meaning at about the same rate (56%) as did humans (50%). Even more striking, the AI system tended to restore the canonical meaning for *the same specific metaphors* as did people (point-biserial correlation = 0.79, $p < .001$, across the 55 individual reversed metaphors). These findings suggest that in interpreting metaphors, GPT-4 resembles humans in its sensitivity to pragmatic constraints on communication.

## 5. Interpreting novel English poems

*5.1 Overview*

A yet more challenging task is to interpret entire poems, which often include metaphors at multiple levels (e.g., an entire poem may itself be a metaphor), coupled with additional poetic devices involving the interplay of symbolic meaning (e.g., synecdoche, metonymy) and sound (e.g., meter, rhyme, alliteration). We asked GPT-4 to interpret eight short English poems that had not been published at the time the system's training was completed (six have never been published at all). A human literary critic (uninformed about the involvement of an AI system) then evaluated the accuracy and quality of these interpretations. We also asked the critic to evaluate two poems generated by GPT-4.

*5.2 Method*

*5.2.1 Selection of poems*

We obtained access to eight short poems (10-29 lines in length), six unpublished and two published since 2022. Three poems were written by each of two contemporary authors, Koon Woon (author of two poetry books, one of which won a PEN Oakland/Josephine Miles Literary Award) and Keith Holyoak (author of four poetry books, and an author of this paper), one poem by Amanda Gorman (former National Youth Poet Laureate and author of multiple poetry books), and one by Jay Parini (author of numerous books spanning poetry, novels, biography, screenplays and criticism). The selected poems (see Section 3 of Supplemental Online Materials) vary in content and style, including both free and formal verse; all incorporate some form of metaphor as well as additional literary devices.

*5.2.2 Assessments of GPT-4's interpretations by a literary critic*

We first gave each poem to GPT-4 without providing the title or author and asked the system to identify it. In each case GPT-4 reported that it was unable to identify the author or source of the poem (as would be expected given that none of the poems were available on the internet when GPT-4 was trained). For each, we then asked GPT-4 to "please give your interpretation of the poem." Each poem was presented in a new chat.

One of the authors of poems that we used (Koon Woon, who also has experience as an editor and publisher of both a poetry journal and an independent literary press) agreed to provide assessments of these interpretations. This literary critic was blind to the origin of the interpretations and was not told that an AI system was involved. The critic first assessed the interpretations provided for the three poems he himself had authored, and then for the remaining poems. We asked him to "take on the role of literary critic or English professor." He was asked to address "whether the interpretation captures the essence of what you see as the meaning of the poem," and to briefly "indicate whatever you think is a major strength or major weakness of the interpretation." Finally, the critic was asked to give each interpretation a letter grade, using the scale: A (excellent), B (good), C (adequate), D (deficient), E (poor). All responses by GPT-4 and by the critic are provided in Section 4 of Supplemental Online Materials.

As a very preliminary test of the literary quality of poems generated by GPT-4, we also asked the human critic to assess the quality of two AI-generated poems (see Section 4 of Supplemental Online Materials). The critic completed this additional task after evaluating the first seven of GPT-4's interpretations of poems (prior to expressing suspicion that an AI system was involved.) His assessments involved comparisons of two human-authored poems with corresponding poems on the same theme generated by GPT-4. One pair of poems was based on "A Skimming Stone" by Holyoak, and another on "Metaphors" by Sylvia Plath. In each case, we first asked GPT-4 to interpret the original poem, and then used its global statement of the theme as a prompt to write its own poem of approximately the same length. The human critic then evaluated the relative strengths and weaknesses of the original poem and the one generated by GPT-4.

*5.3 Results*

The human critic assigned an overall assessment of "excellent" to the first seven interpretations he evaluated, and "good" to the eighth and final one. For all interpretations, the human critic stated that it accurately expressed the symbolic meaning of the poem (e.g., "The critic of this poem gets it"). (For all poems, GPT-4's interpretations, and assessments by the human critic, see Section 3 of Supplemental Online Materials.)

For seven of the eight poems (three different authors, two blind to the involvement of AI), we were also able to obtain the author's assessment of GPT-4's interpretation of their own poem(s). All authors reported the interpretations to be highly accurate. For example, GPT-4 correctly pointed out that the poem "A Skimming Stone" employs the metaphor of a skimming stone to reflect on life, time, and mortality. Moreover, in response to a follow-up query, "Would you say the speaker of the poem is analogous to the child, or to the skimming stone?" GPT-4 reported that "the speaker shifts from being the observer (the child) to being the observed (the stone). This transition signifies the realization of the speaker's own journey through life, mirroring the stone's trajectory." This dynamic metaphoric shift as described by GPT-4 is fully consistent with the poet's intended meaning.

Although the human critic attested to the accuracy of GPT-4's interpretations, he also expressed important reservations related to emotional sensitivity. After finishing all eight assessments, the human critic spontaneously provided an overall characterization of the "critic" being evaluated (i.e., GPT-4): "The critic in all these interpretations zones in on the themes and meanings and their interconnections, but he is weak on evocation portraying just exactly how the reader feels the poem…. This suggests to me that the 'critic' relies on a formula and perhaps is an AI program. It just seems to be without any flair at all even though as far as the straightforward features of a poem goes, it is all correct. It is like a meal that looks done right, even beautifully appealing, but without zesty taste."

In addition to assessing GPT-4's interpretations, we also asked the human critic to evaluate the quality of two poems generated by GPT-4. His evaluations of GPT-4's literary efforts can be characterized as scathing (see Section 4 of Supplemental Online Materials). For example, the human critic describes GPT-4's variant of one poem as "happy rhyming akin to Sunday school joy." Lack of emotional depth, a perceived weakness of GPT-4's otherwise excellent interpretations of human poems, may pose a serious impediment to its capacity to generate novel poems with any literary merit.

## 6. General Discussion

*6.1 Summary*

The language abilities of a state-of-the-art large language model, GPT-4, extend to the interpretation of metaphors, the most prominent form of figurative language. In Study 1 we compared GPT-4 and state-of-the-art language models on a standardized metaphor comprehension task (Liu et al., 2022), and found that GPT-4's performance was substantially more accurate than that achieved by earlier AI models. In Study 2 we compared GPT-4 with humans using a challenging set of novel literary metaphors, generated by Serbian poets and translated into English. Regardless of whether one-off versions of the metaphors had made their way online and potentially into GPT-4's training corpus, the AI system produced metaphor interpretations that human judges (blind to the fact that an AI system was involved in the study) rated as superior to those written by *any* of the human participants—college students at a major public university in the United States. GPT-4 also exhibited a human-like propensity to "make sense" of metaphors presented in a non-canonical form (with source and target reversed). On about half the trials, both people and GPT-4 provided interpretations of reversed metaphors that restored their canonical meaning. Moreover, the AI system resembled humans at the level of individual reversed metaphors, reliably reflecting the probability that a human would restore the canonical meaning at the item level. Our findings add to recent evidence that large language models have begun to acquire some aspects of human pragmatic skills (Barattieri di San Pietro et al., 2023). In Study 3, we found that GPT-4 also produced highly accurate interpretations of novel English poems, exhibiting the ability to analyze larger-scale literary metaphors.

*6.2 Possible computational mechanisms for metaphor interpretation*

Although this study has established that GPT-4 can generate very sensible and human-like interpretations of novel literary metaphors, it leaves open the more fundamental question—how does it do it? Achieving scientific understanding of the operation of LLMs such as GPT-4 continues to be impeded by the refusal of their corporate owners to provide either a detailed account of their training data or access to the internal representations the systems have acquired. We used the words "interpret" and "interpretation" in our queries to GPT-4, but its responses to these close associates of "intelligence" and "thinking" certainly do not establish that this AI system "thinks" in the same way humans do. The caveat raised by the human literary critic with respect to GPT-4's interpretations of poems—an apparent lack of emotional sensitivity—suggest one possible gap between the system's ability to generate meaningful interpretations and the way a human understands literary texts.

*6.3 Apparent limitations of literary processing by GPT-4*

The ability to interpret novel literary metaphors is a remarkable achievement for AI systems. However, it seems premature to conclude that these systems are at present capable of the full range of literary creativity. It is widely claimed that LLMs can "write poetry", and a previous study (Köbis & Mossink (2021) found that non-experts were unable to reliably distinguish poems written by accomplished poets from those generated by an AI system (GPT-2, after pre-training on sets of human-generated poems). The present study provides only a small qualitative comparison; however, it is suggestive that our human literary critic found GPT-4's efforts at writing (as opposed to interpreting) poems to be of low quality. The ability to compose sentences divided into lines linked by end rhymes does not ensure that the product will constitute "poetry", defined by a notable critic as "figurative language, concentrated so that its form is both expressive and evocative" (Bloom, 2004, p. 1). Although generative AI models can learn to imitate the style of individual creators, their poetic products so far appear to lack emotional depth—a deficiency potentially detectable by a literary expert.

Specifically with respect to metaphor, the ability to *generate interpretations of novel metaphor*s must not be confused with the ability to *generate novel metaphors*. Besides being able to generate interpretations of metaphors, as shown here, LLMs can certainly generate texts in which metaphors appear. However, to the best of our current knowledge, the metaphors an LLM might generate are limited to those that human writers have already formed and planted into texts, thereby making humanity's store of metaphors available to be mined by LLMs (Holyoak, 2019).

Metaphor in GPT-4                                                                      22It remains to be seen whether AI systems will at some point be able to create genuinely novel metaphors, rather than only variations of those we humans have made already. The great writer Jorge Luis Borges thought that truly new metaphors still await discovery, at least by humans. New variations of old metaphors can be very beautiful, he acknowledged, "and only a few critics like myself would take the trouble to say, 'Well, there you have eyes and stars and there you have time and the river over and over again.' The metaphors will strike the imagination. But it may also be given to us—and why not hope for this as well?—it may also be given to us to invent metaphors that do not belong, or that do not yet belong, to accepted patterns" (Borges, 2000, p. 41).

**Acknowledgements**

We thank Koon Woon for serving as our literary critic in Study 3. We are grateful to both Koon Woon and Jay Parini for providing us with unpublished poems for use in this study. We also thank Rita Debbaneh, Jocelyn Leison, and Kaycee Stiemke for scoring metaphor interpretations in Study 2. Preparation of this paper was supported by a Google Faculty Research Award to K.J.H.

**Author Contributions**

Conceptualization, K.J.H.; Methodology, D.S., K.J.H., N.I.; Formal Analysis, N.I.; Investigation, D.S., N.I.; Writing – Original Draft, D.S., K.J.H., N.I.; Writing – Review & Editing, D.S., K.J.H., N.I.; Supervision, K.J.H.; Funding Acquisition, K.J.H.

**Declaration of Interests**

The authors declare no competing interests.